%% file: main.tex
\documentclass[10pt,twocolumn,letterpaper]{article}

\usepackage[final]{iccv}      


\definecolor{iccvblue}{rgb}{0.21,0.49,0.74}
\usepackage[pagebackref,breaklinks,colorlinks,allcolors=iccvblue]{hyperref}
\usepackage{multirow}
\usepackage{array}
\usepackage{colortbl}
\usepackage{adjustbox}
\usepackage{algorithm}
\usepackage{algpseudocode}
\usepackage{etoolbox}
\makeatletter
\pretocmd{\@maketitle}{\pagestyle{empty}}{}{}
\makeatother

\usepackage{lineno}
\linenumbers 
\nolinenumbers

\def\paperID{12598} 
\def\confName{ICCV}
\def\confYear{2025}

\title{Model Reveals What to Cache: Profiling-Based Feature Reuse for Video Diffusion Models }

\author{
Xuran Ma\textsuperscript{1,2*} \quad
Yexin Liu\textsuperscript{1,2*} \quad
Yaofu Liu\textsuperscript{2} \quad
Xianfeng Wu\textsuperscript{1,2} \quad
Mingzhe Zheng\textsuperscript{1,2} \quad
Zihao Wang\textsuperscript{1,2} \\
Ser-Nam Lim\textsuperscript{2,3$\dagger$} \quad
Harry Yang\textsuperscript{1,2$\dagger$} \\
\\
$^1$Hong Kong University of Science and Technology \quad
$^2$Everlyn AI \quad
$^3$University of Central Florida \\
\noindent\textsuperscript{*}Equal Contribution \quad
\textsuperscript{$\dagger$}Corresponding Author
\\
{\tt\small xmacb@connect.ust.hk, yliu292@connect.ust.hk, sernam@gmail.com, harryyang.hk@gmail.com} \\
{\small GitHub:~\url{https://github.com/GeekGuru123/ProfilingDiT/tree/main}}
}
\begin{document}

\twocolumn[{
\renewcommand\twocolumn[1][t!]{#1}%
\maketitle
\vspace{-30pt}
\begin{center}
    \centering
    \includegraphics[width=0.95\textwidth]{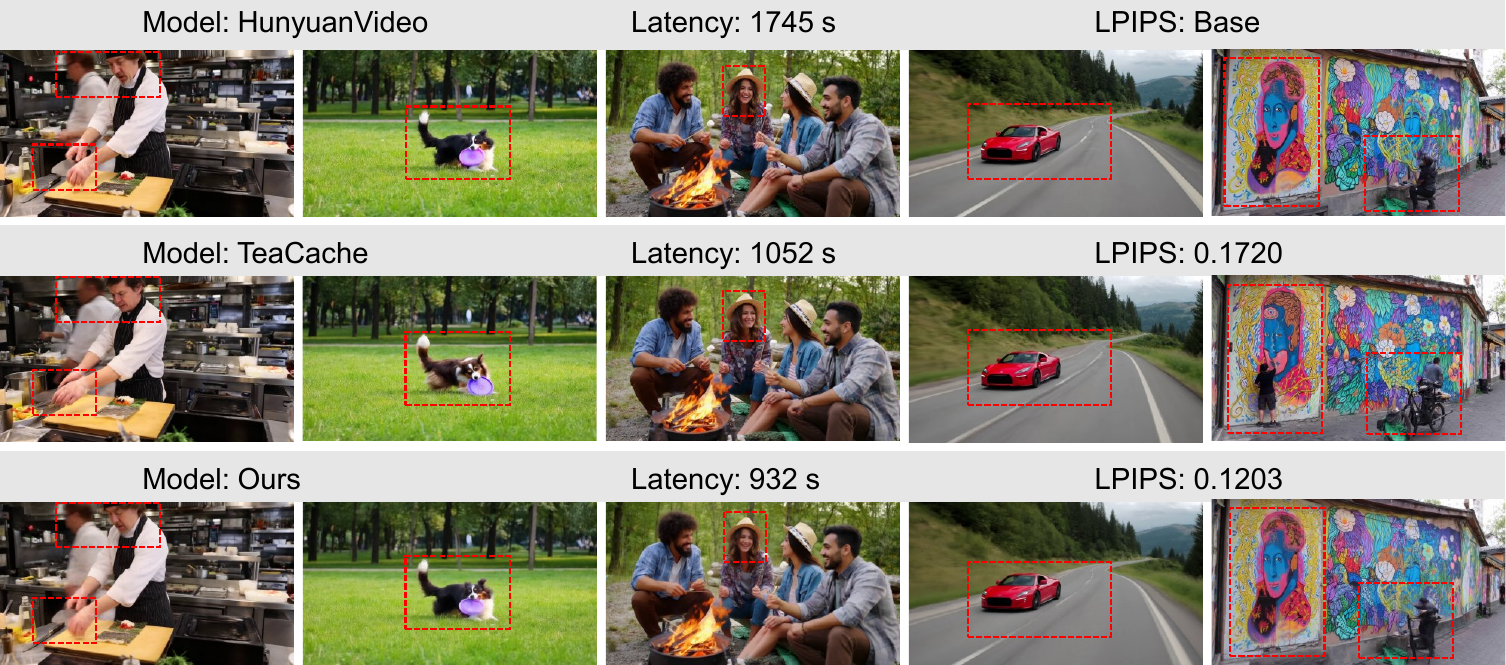}
    \captionof{figure}{Comparison of visual result quality across different methods. The video has a resolution of 780p and consists of 129 frames, with one representative frame extracted from each video for visualization. Our method consistently outperforms TeaCache~\cite{kong2024hunyuanvideo} in both visual quality and efficiency. Latency is evaluated using a single H200 GPU. All results are generated with seed 42.}
    \label{fig:fig1}
\end{center}}
]

\input{sec/0_abstract}

\input{sec/1_intro}
\input{sec/2_related}
\input{sec/3_method}
\input{sec/4_experi}
\input{sec/5_conclusion}
{
    \small
    \bibliographystyle{ieeenat_fullname}
    \bibliography{main}
}


\end{document}

%% file: sec/0_abstract.tex
\begin{abstract}
Recent advances in diffusion models have demonstrated remarkable capabilities in video generation. 
However, the computational intensity remains a significant challenge for practical applications. 
While feature caching has been proposed to reduce the computational burden of diffusion models, existing methods typically overlook the heterogeneous significance of individual blocks, resulting in suboptimal reuse and degraded output quality.
To this end, we address this gap by introducing \textbf{ProfilingDiT}, a novel adaptive caching strategy that explicitly disentangles foreground and background-focused blocks.
Through a systematic analysis of attention distributions in diffusion models, we reveal a \textit{key observation}: 1) Most layers exhibit a consistent preference for either foreground or background regions. 2) Predicted noise shows low inter-step similarity initially, which stabilizes as denoising progresses.
This finding inspires us to formulate a selective caching strategy that preserves full computation for dynamic foreground elements while efficiently caching static background features. 
Our approach substantially reduces computational overhead while preserving visual fidelity. 
Extensive experiments demonstrate that our framework achieves significant acceleration (e.g., 2.01× speedup for Wan2.1) while maintaining visual fidelity across comprehensive quality metrics, establishing a viable method for efficient video generation.
\end{abstract}
\noindent 

%% file: sec/1_intro.tex
\section{Introduction}
Video generation has emerged as a pivotal technology in computer vision, driving advancements across diverse domains from multimedia applications and entertainment to interactive systems~\cite{polyak2025moviegencastmedia,kong2024hunyuanvideo,chen2025goku,zhang2024virbo,pang2024dreamdance,liu2024mycloth,agarwal2025cosmos}. This surge in demand has driven rapid progress in generative modeling~\cite{wu2025lightgen,chen2025temporal,singer2022make}, particularly diffusion-based approaches. Among these methods, the Diffusion Transformer (DiT) architecture~\cite{peebles2023scalable} has gained prominence for its ability to synthesize high-fidelity and temporally consistent videos, supporting downstream tasks including augmented reality~\cite{dong2025enhancing}, automated content creation~\cite{zhang2022edge,xing2025motioncanvas,shao2025finephys}, and AI-driven storytelling~\cite{zheng2024videogen,he2023animate,zhao2024moviedreamer}. However, despite their remarkable success in visual quality and diversity, DiT models face a critical challenge: their inherently sequential denoising process necessitates iterative computations, resulting in substantial computational overhead and substantial memory demands, particularly for high-resolution video generation and large-scale models.




To address this computational bottleneck, researchers have explored various acceleration strategies, including model distillation~\cite{li2023snapfusiontexttoimagediffusionmodel,lin2025diffusionadversarialposttrainingonestep}, quantization~\cite{qdit}, and caching methods. Among these approaches, caching methods have shown particular promise due to their training-free nature and compatibility with pretrained models. Recent advances in this direction include feature reuse mechanisms~\cite{tea}, attention-based caching~\cite{pab,tgate}, and adaptive routing strategies~\cite{adaca,fora}. However, these methods predominantly rely on input-output similarity metrics for cache decisions, overlooking the heterogeneous importance of different blocks. As a result, caching blocks with high similarity may still lead to significant information loss if those blocks play a critical role in semantic modeling. Through systematic analysis of the DiT architecture, we reveal two critical insights: (1) Most layers exhibit a consistent preference for either foreground or background regions. (2) In the initial denoising steps, the predicted noise exhibits low inter-step similarity, gradually stabilizes in the later stages.



Motivated by these findings, we introduce \textbf{ProfilingDiT}, a novel adaptive caching framework that explicitly disentangles foreground and background-focused computations. Our approach leverages the observed semantic separation in transformer blocks to implement a dual-granularity caching strategy. At the block level, we selectively cache features from layers that predominantly attend to background regions, while preserving full computation for those focused on foreground details. At the step level, we analyze the L1 distance between consecutive denoising steps as a proxy for prediction stability. In the early steps, the distance is high, reflecting rapid semantic transitions. Thus, caching is disabled to retain full update precision. As diffusion progresses, the distance stabilizes, allowing efficient caching at larger intervals. In the final steps, the distance increases again, indicating the need for finer-grained updates and a reduced caching rate.

Our comprehensive experiments demonstrate the effectiveness of ProfilingDiT. For instance, on the Wan2.1~\cite{wan2.1} model, our approach achieves a 2.01× speed-up in inference while maintaining a competitive LPIPS of 0.1256. In addition to perceptual quality, we also observe substantial gains in other metrics, with PSNR improved to 22.02 and SSIM reaching 0.7899. These results confirm that ProfilingDiT enables low-latency generation without compromising visual fidelity.


The primary contributions of our work include:
\begin{itemize}
\item \textbf{Attention Analysis of DiT in diffusion:} We investigate the attention distributions and noise dynamics in DiT during the denoise process, revealing consistent foreground-background preferences across blocks and temporally evolving noise similarity across diffusion steps.
\item \textbf{Adaptive Caching Framework:} We introduce a novel dual-granularity caching strategy that leverages semantic separation for optimal computational resource allocation, significantly advancing the state-of-the-art in efficient video generation.
\item \textbf{Empirical Validation:} Through extensive experimentation, we demonstrate that ProfilingDiT achieves substantial computational efficiency (up to 2.01× acceleration) while maintaining high-quality video generation across diverse evaluation metrics.
\end{itemize}

%% file: sec/2_related.tex
\section{Related Works}
\paragraph{Video Diffusion Model.}
Diffusion models~\cite{song2022denoisingdiffusionimplicitmodels,ho2020denoisingdiffusionprobabilisticmodels,lu2022dpmsolverfastodesolver,chen2025temporal} have emerged as foundational frameworks for generative tasks, showcasing remarkable capabilities in producing diverse and high-quality outputs. Initially employing U-Net architectures, these models have been highly successful in both image and video generation tasks. However, the inherent scalability limitations of U-Net-based diffusion models~\cite{feng2024fancyvideo,wang2024loopanimate} restrict their effectiveness for complex, large-scale generation tasks. To overcome this constraint, DiT was introduced, effectively leveraging the scalability and representational power of transformer architectures. Notably, models like Sora have demonstrated significant advancements by employing transformer-based diffusion frameworks to capture complex dynamics of real-world phenomena. Recent research has further evolved video diffusion models from initial 2D architectures with additional temporal modeling~\cite{opensora,hong2022cogvideo} to fully 3D diffusion frameworks~\cite{osp,yang2024cogvideox,wan2.1,kong2024hunyuanvideo,stepvideo}. This architectural shift enables better handling of spatial-temporal dependencies, significantly improving video quality, motion coherence, and generation fidelity over extended sequences.

 \begin{figure*}[t!]
    \centering
    \includegraphics[width=0.9\textwidth]{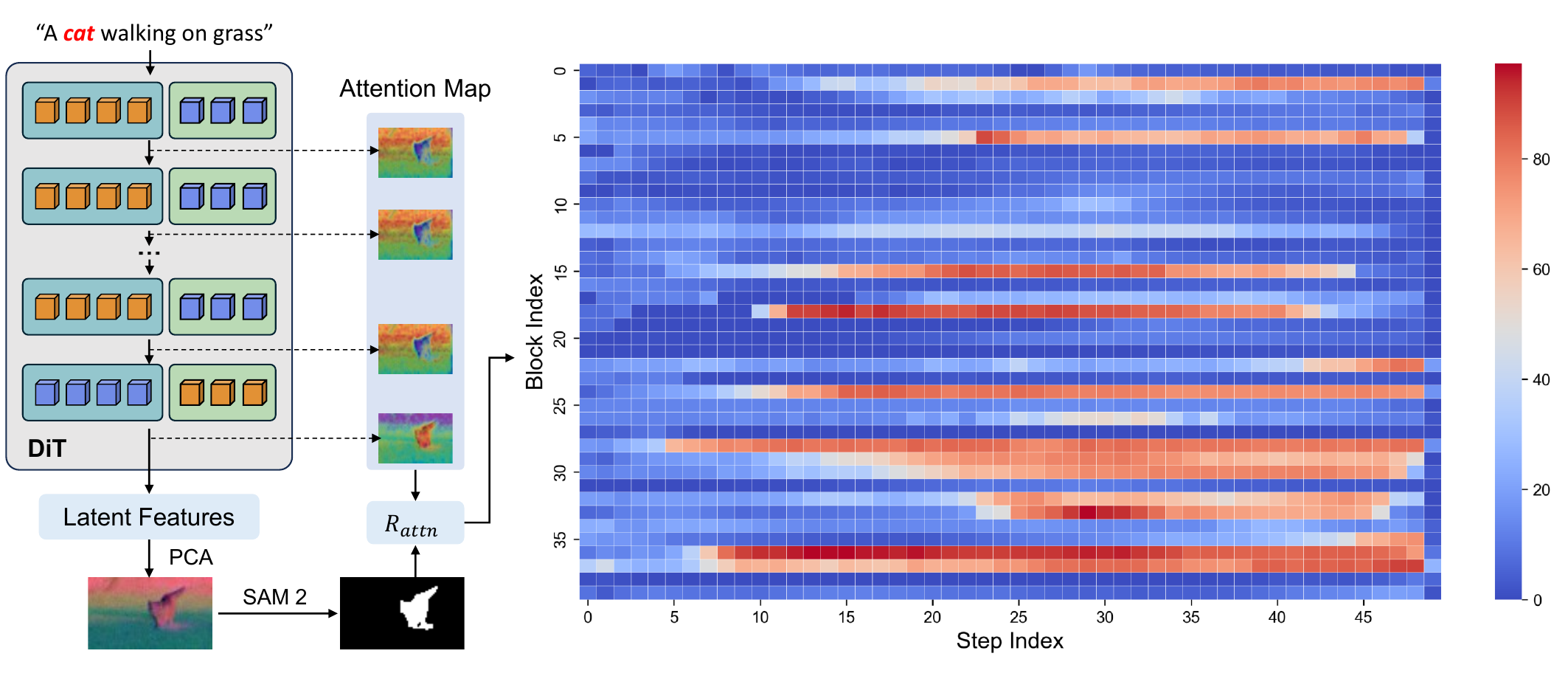} 
    \caption{\textbf{Visualization of attention focus across blocks and diffusion steps in DiT.}
  \textit{Left}: We compute the foreground attention ratio \( R_{\text{attn}} \) for each block using attention maps and semantic masks derived from SAM2. 
  \textit{Right}: Heatmap of \( R_{\text{attn}} \) across 40 blocks and 50 denoising steps in HunyuanVideo~\cite{kong2024hunyuanvideo}. Warmer colors indicate higher foreground focus. Details are in Sec.~\ref{Attention Pattern Analysis}.}
    \label{fig: Pipeline of analysis}
\end{figure*}

\paragraph{Caching Methods for Video Diffusion Transformers.}
Accelerating video diffusion model inference remains an active area of research, with strategies such as distillation, quantization, and caching being explored extensively. Distillation approaches~\cite{li2023snapfusiontexttoimagediffusionmodel,lin2025diffusionadversarialposttrainingonestep} typically aim to compress larger models into smaller, efficient versions, whereas quantization methods~\cite{qdit} reduce computational overhead by lowering numerical precision. In contrast, caching methods primarily focus on feature reuse during inference to minimize redundant computations. Existing caching approaches for Diffusion Transformers have made notable strides in enhancing inference speed. For instance, TeaCache~\cite{tea} reuses noise features based on a routing decision mechanism. Methods like PAB~\cite{pab} and TGate~\cite{tgate} specifically reuse attention outputs by identifying critical attention components. Delta~\cite{delta} focuses on reusing block outputs, while AdaCache and Fora~\cite{adaca,fora} introduce adaptive routers to optimize reuse across attention and MLP layers. Other advancements include FasterCache~\cite{fastercache}, which proposes a classifier-free guidance (CFG) caching strategy, and ToCa~\cite{toca}, introducing token-level reuse and compression. However, the transition to fully 3D transformer architectures significantly increases memory usage, making attention-based caching less viable and consequently positioning block-level and noise reuse as preferable methods for managing computational resources in video diffusion models. 

%% file: sec/3_method.tex
\section{Methodology}
\subsection{Preliminaries}
\paragraph{Denoising Diffusion Models.}
Diffusion models simulate visual generation through a sequence of iterative denoising steps. Starting from random noise, these models progressively refine the noise until it closely approximate samples from the desired distribution. During the forward diffusion process, Gaussian noise is incrementally added over 
\( T \) steps to a data point \( x_0 \) sampled from the real distribution \( q(x) \):
\begin{equation}
    x_t = \sqrt{\alpha_t} x_{t-1} + \sqrt{1 - \alpha_t} z_t, \quad t = 1, \dots, T,
\end{equation}
where 
\( \alpha_t \in [0,1] \) determines the noise intensity, and \( z_t \sim \mathcal{N}(0, I) \) represents Gaussian noise. As \( t \) increases, \( x_t \) becomes progressively noisier, ultimately resembling a normal distribution \( \mathcal{N}(0, I) \) when \( t = T \)
. The reverse diffusion process is designed to reconstruct the original data from its noisy counterpart:
\begin{equation}
    p_{\theta}(x_{t-1} | x_t) = \mathcal{N}(x_{t-1}; \mu_{\theta}(x_t, t), \Sigma_{\theta}(x_t, t)),
\end{equation}
where 
\( \mu_{\theta} \) and \( \Sigma_{\theta} \)
 are learned parameters defining the mean and covariance.

\begin{figure}[t!]
    \centering
    \includegraphics[width=0.5\textwidth]{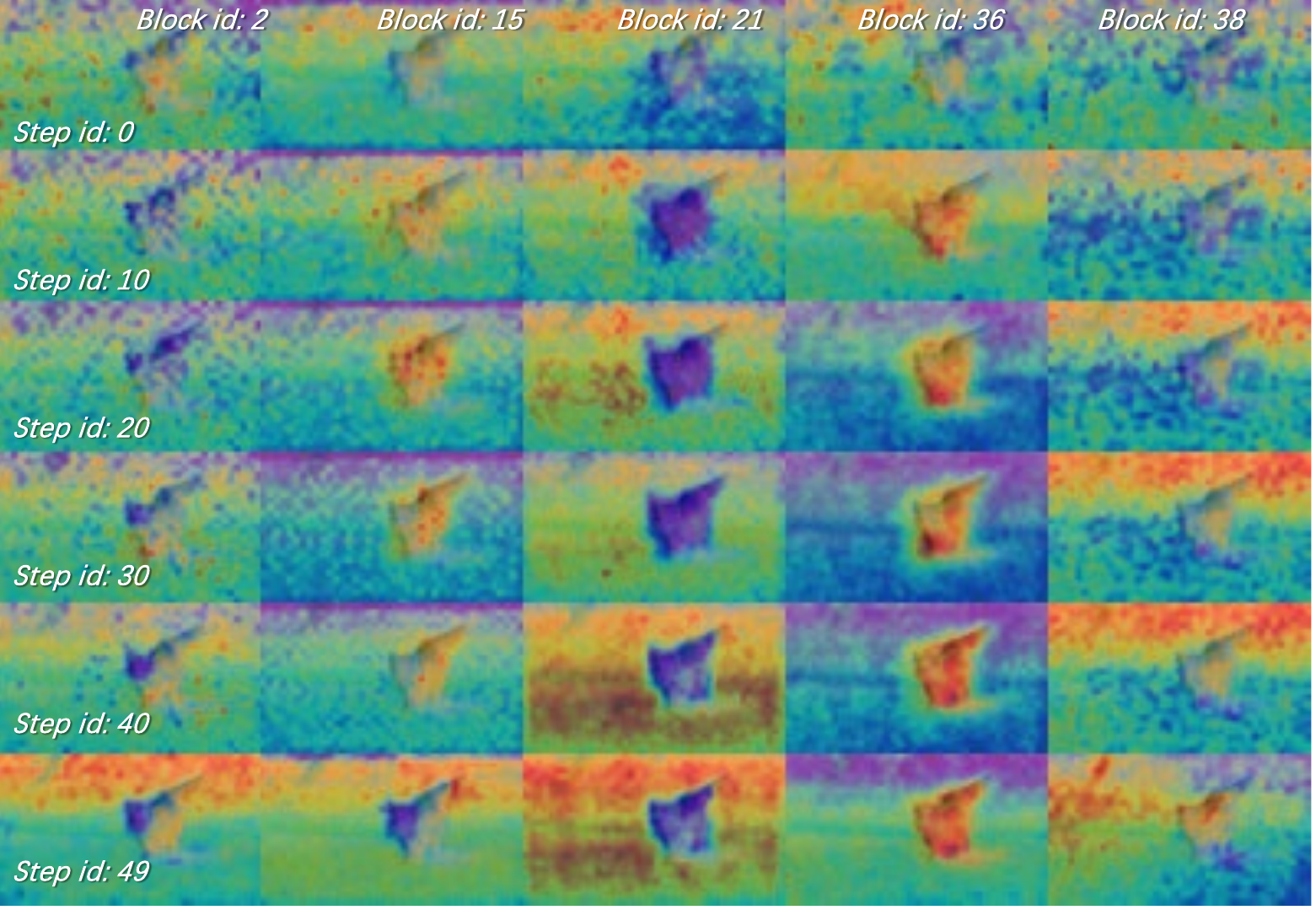} 
    \caption{Visualization of the attention scores heatmap. Attention scores are displayed at the block level (rows) and step level (columns), revealing consistent patterns across steps. Red denotes regions with the highest attention, followed by green and blue, while purple indicates areas with the lowest attention.}
    \label{fig: Heatmap}
\end{figure}

\paragraph{Attention Mechanism.}
The attention mechanism in DiT operates by computing Query (\(\mathbf{Q}\)), Key (\(\mathbf{K}\)), and Value (\(\mathbf{V}\)) matrices from the input latent features \(\mathbf{X}\). These matrices are obtained through learned linear projections:
\begin{equation}
    \mathbf{Q} = \mathbf{X} \mathbf{W}^Q, \quad \mathbf{K} = \mathbf{X} \mathbf{W}^K, \quad \mathbf{V} = \mathbf{X} \mathbf{W}^V
\end{equation}
where \(\mathbf{W}^Q\), \(\mathbf{W}^K\), and \(\mathbf{W}^V\) are the projection matrices for queries, keys, and values, respectively. The attention scores are computed as follows:
\begin{equation}
    \mathbf{A} = \text{softmax}\left(\frac{\mathbf{Q} \mathbf{K}^\top}{\sqrt{d_k}}\right)
\end{equation}
where \(\mathbf{A}\) is the attention matrix, with \( A_{ij} \) representing the attention score of token \( j \) attending to token \( i \).

\subsection{Attention Pattern Analysis}
\label{Attention Pattern Analysis}
Given an input prompt, such as \textit{``a cat walking on grass"}, we analyze the attention behavior within DiT blocks. Each block contains a self-attention layer, from which token embeddings \( \mathbf{X} \in \mathbb{R}^{B \times N \times C} \) are extracted across all denoising steps. The token dimension \( N \) is divided into \( T \) segments, each corresponding to one video frame. Attention scores are computed for each frame by summing across rows in the attention matrix.

\textbf{Block-Level Analysis.} To determine whether each block predominantly attends to foreground or background regions, we first extract the noise prediction at each step (shape: \( B \times C \times T \times H \times W \)). To segment the latent space into foreground and background, we apply PCA~\cite{abdi2010principal} to reduce the channel dimension to 3, forming RGB-like visualizations as shown in Fig.~\ref{fig: Pipeline of analysis}. We then use SAM2~\cite{ravi2024sam} to generate binary masks of the foreground.

Tokens with attention scores exceeding a predefined threshold are selected.  Given a transformer block's attention matrix $\mathbf{A} \in \mathbb{R}^{N \times N}$, where $N$ denotes the number of spatial-temporal tokens, we compute the aggregated attention score for each token $i$ as:
\begin{equation}
\bar{a}_i = \frac{1}{N} \sum_{j=1}^{N} A_{ij}
\end{equation}

The proportion of these high-attention tokens that lie within the foreground mask is computed. Although a single frame is shown for illustration, the calculation is performed over all frames in a block and averaged to yield a score \(R_{attn}\), reflecting the concentration of high-attention tokens within foreground regions.

\begin{figure}[t!]
    \centering
    \includegraphics[width=0.5\textwidth]{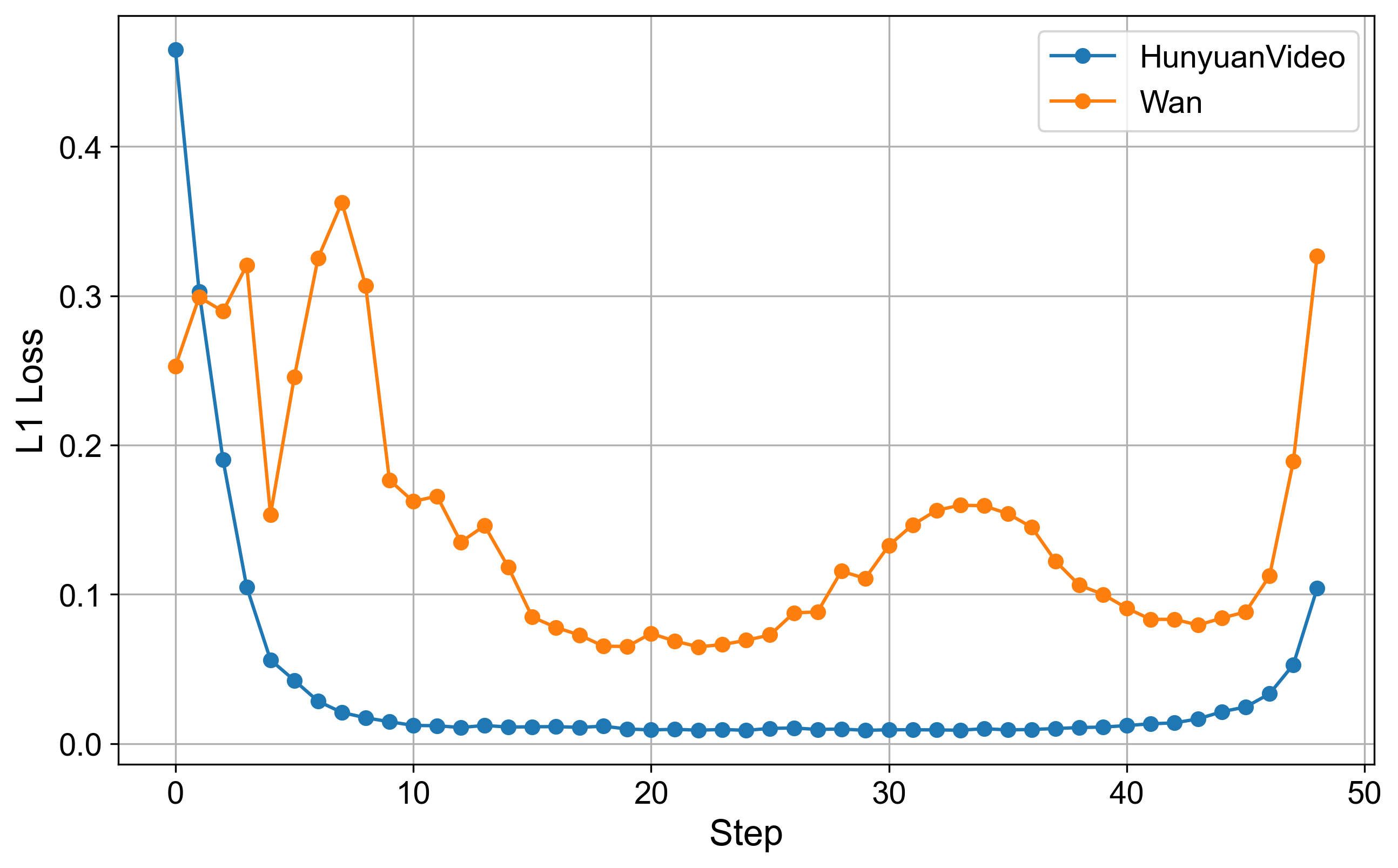} 
    \caption{L1 distance of predicted noise on different steps.}
    \label{fig: attention step analysis}
\end{figure}

\begin{figure*}
    \centering
    \includegraphics[width=\textwidth]{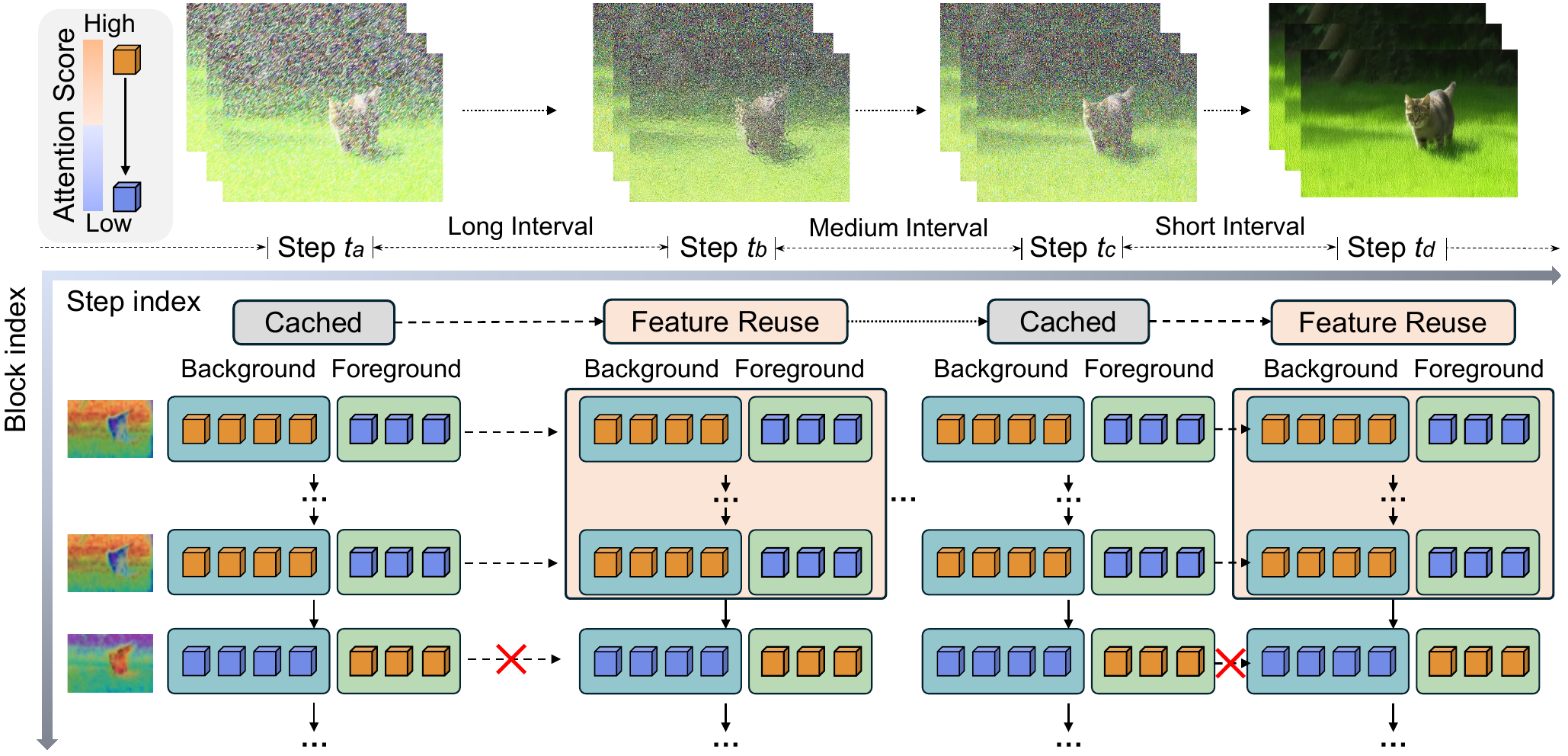} 
    \caption{Pipeline of ProfilingDiT: Tokens highlighted in yellow and blue indicate higher and lower attention scores, respectively. The alignment between attention regions and object regions determines whether the corresponding block outputs should be cached.}
    \label{fig: ProfilingDiT pipeline}
\end{figure*}

\begin{equation}
R_{attn} = \frac{N_{\text{high} \cap \text{fg}}}{N_{\text{fg}}}
\end{equation}
where \( N_{\text{high} \cap \text{fg}} \) denotes the number of tokens that overlap between high-attention\((\bar{a}_i > \text{threshold})\)
 tokens and the foreground mask, and \( N_{\text{fg}} \) is the total number of foreground tokens.
Then we can get a block list from:
We define the foreground ($\mathcal{F}$) and background ($\mathcal{B}$) blocks list based on a predefined threshold $\tau$:
\begin{equation}
\mathcal{F} = [ i \mid \ R_{attn} \geq \tau ], \quad \mathcal{B} = [ i \mid \ R_{attn} < \tau ]
\end{equation}

Fig.~\ref{fig: Pipeline of analysis} and Fig.~\ref{fig: Heatmap} visualize the score \(R_{attn}\) of 40 single blocks of MMDiT within the HunyuanVideo~\cite{kong2024hunyuanvideo} on all 50 steps. We observe that \(R_{attn}\) for each block remains consistent across steps, supporting the hypothesis that blocks exhibit an intrinsic focus on either foreground or background. This attention preference remains stable throughout the denoising process. The attention distribution provides insights into how each block prioritizes different regions of the input. Specifically, some blocks consistently focus on foreground content, while others emphasize the background. Notably, early steps are more background-oriented, with a gradual shift toward foreground attention after step 6.

\textbf{Step-Level Analysis.} We extract noise predictions at different steps from HunyuanVideo~\cite{kong2024hunyuanvideo} and Wan2.1~\cite{wan2.1} and compute step-level similarity. Fig.~\ref{fig: attention step analysis} shows the L1 distance between consecutive noise predictions during denoising steps. In the early steps, the L1 distance is high, indicating rapid changes in the predictions. Accordingly, no caching is applied to allow precise updates. In later steps, the L1 distance stabilizes, enabling efficient caching with larger intervals. In the final steps, the L1 distance increases, implying less caching rate.

\subsection{ProfilingDiT}
\label{ProfilingDiT}

We introduce ProfilingDiT, a semantically aware adaptive caching framework that optimizes computational efficiency by explicitly distinguishing between foreground and background processing in diffusion transformers. As shown in Fig.~\ref{fig: ProfilingDiT pipeline}, our approach implements a dual-granularity caching strategy, operating at both block and temporal levels to maximize computational resource utilization while preserving generation quality.

\begin{table*}[t!]
    \centering
    \begin{tabular}{l|ccccc|cc}
        \toprule
        \multirow{2}{*}{\textbf{Method}} & \multicolumn{5}{c|}{\textbf{Visual Quality Evaluation}} & \multicolumn{2}{c}{\textbf{Efficiency Evaluation}} \\
        \cmidrule(lr){2-6} \cmidrule(lr){7-8}
        & VBench $\uparrow$ & LPIPS $\downarrow$ & PSNR $\uparrow$ & SSIM $\uparrow$ & FID $\downarrow$ & Latency (s) $\downarrow$& Speedup $\uparrow$ \\
        \midrule
        HunyuanVideo (720P, 129frames)~\cite{kong2024hunyuanvideo} & 0.7703 & --    & --    & --    & --    & 1745 & -- \\
        TeaCache (slow)~\cite{tea}               & \textbf{0.7700} & \underline{0.1720} & \underline{21.91} & \underline{0.7456} & \underline{77.67} & 1052 & 1.66$\times$ \\
        TeaCache (fast)~\cite{tea}               & \underline{0.7677} & 0.1830 & 21.60 & 0.7323 & 83.85 & \textbf{753} & \textbf{2.31$\times$} \\
        \textit{Ours}                            & 0.7642 & \textbf{0.1203} & \textbf{26.44} & \textbf{0.8445} & \textbf{41.10} & \underline{932} & \underline{1.87$\times$} \\
        \bottomrule
    \end{tabular}
    \caption{Quantitative comparison with prior methods under HunyuanVideo baselines. ``\textit{Ours}'' consistently outperforms others in visual quality while maintaining strong efficiency. $\uparrow$: higher is better; $\downarrow$: lower is better. Bolding and underlining indicate the best and second-best performance, respectively.}
    \label{tab: compare 1}
\end{table*}

\begin{table*}[t!]
    \centering
    \begin{tabular}{l|ccccc|cc}
        \toprule
        \multirow{2}{*}{\textbf{Method}} & \multicolumn{5}{c|}{\textbf{Visual Quality Evaluation}} & \multicolumn{2}{c}{\textbf{Efficiency Evaluation}} \\
        \cmidrule(lr){2-6} \cmidrule(lr){7-8}
        & VBench $\uparrow$ & LPIPS $\downarrow$ & PSNR $\uparrow$ & SSIM $\uparrow$ & FID $\downarrow$ & Latency (s) $\downarrow$& Speedup $\uparrow$ \\
        \midrule
        Wan2.1 (480P, 81frames)~\cite{wan2.1}                     & 0.7582 & --    & --    & --    & --    & 497 & -- \\
        TeaCache (fast)~\cite{tea}               & \underline{0.7604} & \underline{0.2913} & \underline{16.17} & \underline{0.5685} & \underline{117.61} & \underline{249} & \underline{2.00$\times$} \\
        \textit{Ours}                            & \textbf{0.7615} & \textbf{0.1256} & \textbf{22.02} & \textbf{0.7899} & \textbf{62.56} & \textbf{247} & \textbf{2.01$\times$} \\
        \bottomrule
    \end{tabular}
    \caption{Quantitative comparison with prior methods under Wan2.1 baselines. Bolding and underlining indicate the best and second-best performance, respectively.}
    \label{tab: compare 2}
\end{table*}


\noindent\textbf{Block-Level Caching.} Motivated by Delta-DiT~\cite{delta}, we cache block-level deviations between the output and input hidden states.
Specifically, for a block $i$, we compute:
\begin{equation}
\Delta_{\text{block}} = 
\begin{cases}
h_{\text{out}}^{(i)} - h_{\text{in}}^{(i)}, & \text{if } i \in \mathcal{B}, \\
0, & \text{if } i \in \mathcal{F}.
\end{cases}
\end{equation}

\begin{figure}[t!]
    \centering
    \includegraphics[width=0.5\textwidth]{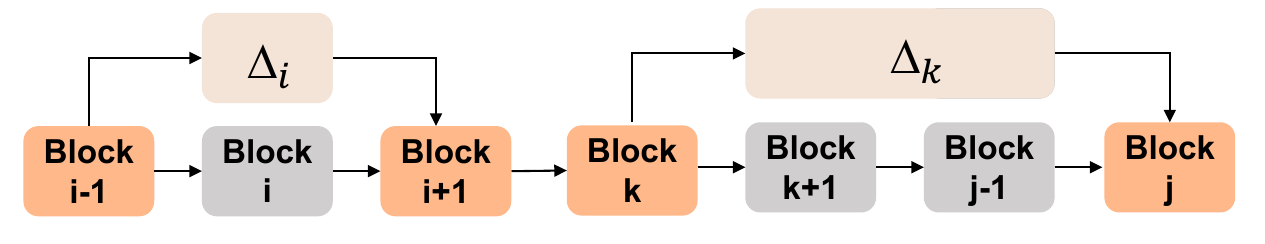} 
    \caption{Details of delta list scheme in Sec.\ref{ProfilingDiT}.}
    \vspace{-8pt}
    \label{fig: attention step analysis pipeline}
\end{figure}

However, based on our previous observations, the block focuses on the background list that contains both consecutive numbers and individual numbers. The Delta-DiT only caches a single group of consecutive blocks, but we need to process all the blocks in the list. Therefore, we have designed the following \textit{delta list} scheme, as shown in Fig.~\ref{fig: attention step analysis pipeline}.

\begin{equation}
\Delta_i, \Delta_k, \dots = 
\begin{cases}
h_{\text{out}}^{(i)} - h_{\text{in}}^{(i)}, & \text{if } i \in \mathcal{B} , \\
h_{\text{out}}^{(j)} - h_{\text{in}}^{(k)}, & \text{if } i \in \mathcal{B}, \, j - k > 1 , \\
0, & \text{if } i \in \mathcal{F}.
\end{cases}
\end{equation}
where $j$ and $k$ denote the start and end indices of a continuous interval. and $i$ is an individual number.

To propagate cached features across multiple steps, we maintain an accumulated cache state $\mathcal{C}^{(i)}_t$ at each step $t$, which updates recursively as follows:
\begin{equation}
\mathcal{C}^{(i)}_{t+1} = \mathcal{C}^{(i)}_t + \Delta_{\text{t}}^{(i)}
\end{equation}
where $\mathcal{C}^{(i)}_0 = 0$ for all blocks at the initial step. This can selectively accumulate deviations for background-focused blocks and optimize computation for continuous block sequences, achieving a balance between computational efficiency and generation quality.

\noindent\textbf{Step-Level Caching.} 
Motivated by the trend observed in Fig.~\ref{fig: attention step analysis}, where the \( L_1 \) loss is larger in earlier steps and stabilizes over time. We also introduce a temporal caching strategy that distinguishes between cache computation and cache utilization phases. For a given diffusion step $s$, we define a caching interval $T_s$ that modulates the frequency of cache updates:
\begin{equation}
T_s = f(s),
\end{equation}
where $f(s)$ is a monotonically decreasing function representing the caching interval. For example, the intervals can be $f(s) \in \{12, 9, 6, 3\}$, indicating that caching becomes progressively more frequent over time.


For steps $s$ beyond an initial warm-up phase (i.e., $s > s_0$), we maintain two distinct step sets: cache computation steps $\mathcal{S}_{\text{comp}}$ and cache utilization steps $\mathcal{S}_{\text{use}}$, where:
\begin{equation}
\mathcal{S}_{\text{comp}} = \{s \mid s \bmod T_s = 0, s > s_0\}
\end{equation}

During cache computation steps ($s \in \mathcal{S}_{\text{comp}}$), the hidden state update follows:
\begin{equation}
h_s = h_s + \Delta_{\text{cache}}^{(i)},
\end{equation}
where $\Delta_{\text{cache}}^{(i)}$ is computed according to our block-level strategy. For cache utilization steps ($s \in \mathcal{S}_{\text{use}}$), we directly reuse cached features for background blocks while maintaining full computation for foreground blocks:
\begin{equation}
h_s = 
\begin{cases}
h_s + \Delta_{\text{cache}}^{(i)}, & \text{if } i \in \mathcal{B}, \\
\text{Block}_i(h_s), & \text{if } i \in \mathcal{F}.
\end{cases}
\end{equation}

To effectively balance computational efficiency with generation quality, we implement an adaptive caching interval:
\begin{equation}
T_s = T_{\max} - (T_{\max} - T_{\min}) \cdot \frac{s - s_0}{S - s_0},
\end{equation}
where $T_{\max}$ and $T_{\min}$ represent the initial and final caching intervals, $S$ denotes the total diffusion steps, and $s_0$ indicates the warm-up threshold. This progressive adjustment of caching frequency enables more frequent feature reuse as the diffusion process stabilizes, while preserving computational precision during critical early steps.

%% file: sec/4_experi.tex
\section{Experiment}
\begin{table*}[t!]
    \centering
    \begin{tabular}{l|ccccc|cc}
        \toprule
        \multirow{2}{*}{\textbf{Method}} & \multicolumn{5}{c|}{\textbf{Visual Quality Evaluation}} & \multicolumn{2}{c}{\textbf{Efficiency Evaluation}} \\
        \cmidrule(lr){2-6} \cmidrule(lr){7-8}
        & VBench $\uparrow$ & LPIPS $\downarrow$ & PSNR $\uparrow$ & SSIM $\uparrow$ & FID $\downarrow$ & Latency (s) $\downarrow$ & Speedup $\uparrow$ \\
        \midrule
        HunyuanVideo (Baseline) & 0.7703 & --     & --    & --    & --    & 1745 & -- \\
        \midrule
        Foreground-focused Block Reuse & \textbf{0.7685} & \underline{0.1449} & \underline{24.23} & \underline{0.8065} & 58.53 & 1113 & 1.57$\times$ \\
        Background-focused Block Reuse & \underline{0.7642} & \textbf{0.1203} & \textbf{26.44} & \textbf{0.8445} & \textbf{41.10} & \textbf{932}  & \textbf{1.87$\times$} \\
        \midrule
        Background+Foreground  (\textit{Split}) & 0.7637 & 0.1753 & 23.83 & 0.7964 & 61.79 & \underline{1056} & \underline{1.65$\times$} \\
        Background+Foreground  (\textit{Alternate}) & 0.7623
 & 0.1875 & 23.46 & 0.7893 & \underline{55.89} & \underline{1056} & \underline{1.65$\times$} \\
        \bottomrule
    \end{tabular}
    \caption{Ablation on block-level reuse strategies. Foreground reuse achieves the best perceptual quality and distortion metrics. \textit{``Split”} alternates block types across step groups, while \textit{``Alternate”} switches reuse types between steps. Bolding and underlining indicate the best and second-best performance, respectively.}
    \label{tab: Ablation1}
\end{table*}

\begin{table*}[t!]
    \centering
    \begin{tabular}{l|ccccc|cc}
        \toprule
        \multirow{2}{*}{\textbf{Method}} & \multicolumn{5}{c|}{\textbf{Visual Quality Evaluation}} & \multicolumn{2}{c}{\textbf{Efficiency Evaluation}} \\
        \cmidrule(lr){2-6} \cmidrule(lr){7-8}
        & VBench $\uparrow$ & LPIPS $\downarrow$ & PSNR $\uparrow$ & SSIM $\uparrow$ & FID $\downarrow$ & Latency (s) $\downarrow$& Speedup $\uparrow$ \\
        \midrule
        HunyuanVideo (Baseline) & 0.7703 & --     & --    & --    & --    & 1745 & -- \\
        \midrule
        \multicolumn{8}{l}{\textit{Foreground-focused Block Reuse}} \\
        \quad Foreground + \textit{Stepwise}      & \underline{0.7685} & 0.1449 & 24.23 & 0.8065 & 58.53 & 1113 & 1.57$\times$ \\
        \quad Foreground + \textit{Step Inverse}  & 0.7681 & 0.1785 & 24.80 & 0.7735 & 65.57 & 1098 & 1.59$\times$ \\
        \quad Foreground + \textit{Step Average}  & \textbf{0.7698} & \underline{0.1303} & 25.10 & 0.8193 & \underline{50.29} & 1089 & 1.60$\times$ \\
        \midrule
        \multicolumn{8}{l}{\textit{Background-focused Block Reuse}} \\
        \quad Background + \textit{Stepwise}      & 0.7642 & \textbf{0.1203} & \textbf{26.44} & \textbf{0.8445} & \textbf{41.10} & \underline{932}  & \underline{1.87$\times$} \\
        \quad Background + \textit{Step Inverse}  & 0.7517 & 0.2529 & 24.49 & 0.7495 & \textbf{41.10} & 934  & 1.87$\times$ \\
        \quad Background + \textit{Step Average}  & 0.7627 & 0.1571 & \underline{25.22} & \underline{0.8275} & 53.20 & \textbf{924}  & \textbf{1.89$\times$} \\
        \bottomrule
    \end{tabular}
    \caption{Ablation on step-level reuse strategies across background and foreground blocks. \textit{``Stepwise"} denotes decreasing reuse intervals; \textit{``Step Inverse"} denotes increasing reuse intervals; \textit{``Step Average"} denotes fixed reuse intervals. Bolding and underlining indicate the best and second-best performance, respectively.}
    \label{tab: Ablation0}
    \vspace{-10pt}
\end{table*}

\subsection{ Implementation Details}
Comparison experiments for HunyuanVideo are completed on NVIDIA H200 140GB GPUs using PyTorch. FlashAttention~\cite{dao2022flashattention} is enabled by default for all experiments. The default parameters follow the official HunyuanVideo settings: 720×1280 resolution, 129 frames, 50 inference steps, and a fixed seed. Further experiments on Wan2.1~\cite{wan2.1} is carried on H100 80GB GPUs, the hardware setting between baselines and ours are the same.
\subsection{Metrics}
To evaluate the performance of video synthesis acceleration methods, we focus on two key aspects: inference efficiency and visual quality. For inference efficiency, we use Floating Point Operations (FLOPs) and inference latency as our primary metrics. We employ VBench~\cite{huang2024VBench}, LPIPS~\cite{zhang2018unreasonable}, PSNR, and SSIM for visual quality assessment. VBench~\cite{huang2024VBench} is a comprehensive benchmark suite for video generative models, aligning well with human perception and providing valuable insights from multiple perspectives. Additionally, LPIPS, PSNR, and SSIM measure the similarity between videos generated by the accelerated sampling method and those from the original model: PSNR (Peak Signal-to-Noise Ratio) evaluates pixel-level fidelity, LPIPS (Learned Perceptual Image Patch Similarity) measures perceptual consistency, and SSIM (Structural Similarity Index) assesses structural similarity.

\subsection{Comparison Experiment}
We evaluate our proposed method with three baselines: HunyuanVideo~\cite{kong2024hunyuanvideo}, TeaCache-slow~\cite{tea}, and TeaCache-fast~\cite{tea}, using six performance metrics: VBench, LPIPS, PSNR, SSIM, FID, and latency. 

Tab.~\ref{tab: compare 1} and Tab.~\ref{tab: compare 2} present the results. The HunyuanVideo serves as the baseline with a VBench score of 0.7703 and a latency of 1745 ms.   \textbf{TeaCache-slow} shows a marginal decrease in VBench (0.7700) while achieving moderate visual quality (LPIPS = 0.1720, PSNR = 21.91, SSIM = 0.7456) with a significantly reduced latency of 1052 ms (1.66 faster than the baseline). \textbf{TeaCache-fast} further lowers latency to 753 ms (2.31 faster than the baseline) at the cost of degraded image quality, reflected in higher LPIPS (0.1830) and FID (83.85).
Our approach achieves notable improvements in reconstruction metrics: LPIPS is reduced to 0.1203, PSNR increases to 26.44, SSIM improves to 0.8445, and FID decreases to 41.10, indicating superior perceptual quality compared to the TeaCache variants. Additionally, our method achieves a latency of 932 ms (1.87 faster than the baseline), demonstrating efficiency gains without compromising performance.

\begin{figure*}[t!]
    \centering
    \includegraphics[width=\textwidth]{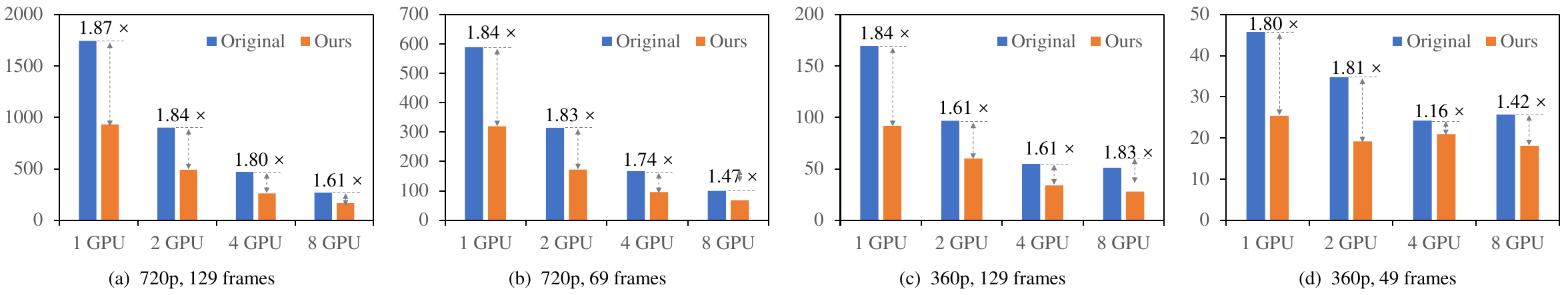} 
    \caption{Inference efficiency of our method at different resolutions and video lengths.}
    \label{fig: resolutions and video lengths}
\end{figure*}

\begin{figure}[t!]
    \centering
    \includegraphics[width=0.5\textwidth]{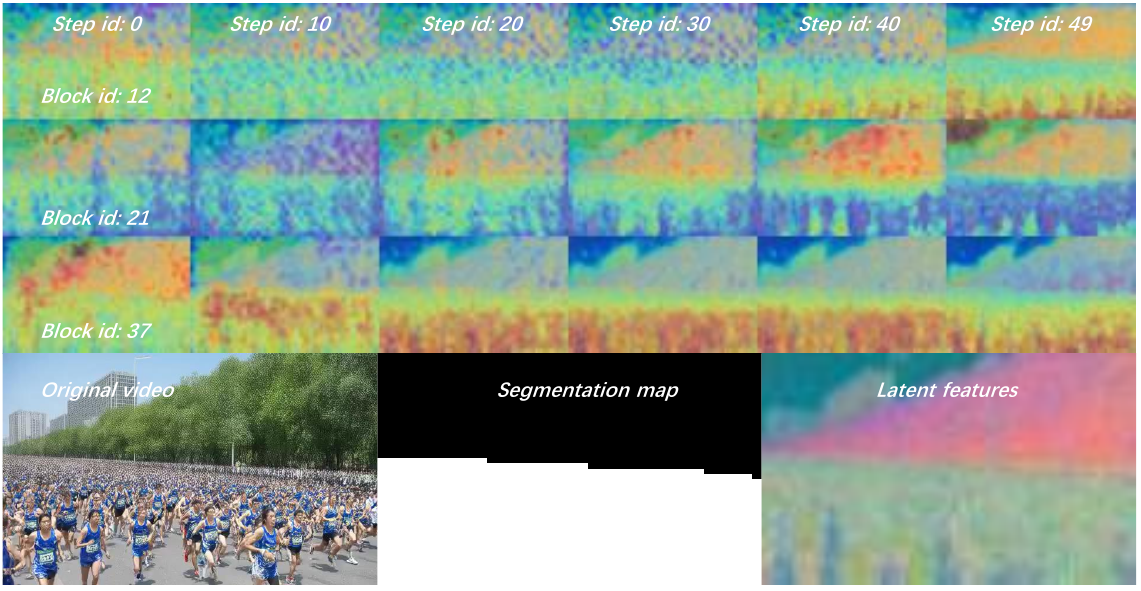} 
    \caption{Attention map visualization for a multi-object video.}\
    \vspace{-15pt}
    \label{fig: multi-onject}
\end{figure}
\subsection{Ablation Study}

We conduct ablation studies to investigate the impact of reusing background and foreground information at different stages on both visual quality and runtime performance.

\paragraph{Ablation on Block-Level Reuse.}
We first investigate the impact of reusing different types of blocks—\textbf{Foreground-focused} versus \textbf{Background-focused}. The results are shown in Tab.~\ref{tab: Ablation1}. Our analysis shows that \textbf{Foreground-focused block reuse} (\ie, reusing blocks that focus on foreground content) yields a competitive VBench score (0.768), but struggles with fine-grained visual quality (LPIPS = 0.1449, PSNR = 24.23, FID = 58.53). In contrast, \textbf{Background-focused block reuse} (\ie, reusing blocks that focus on dynamic background content) is more effective for detailed reconstruction, achieving substantially better LPIPS (0.1203), PSNR (26.44), and SSIM (0.8445), along with a lower FID of 41.10. These results suggest that background features are more spatially localized and temporally variant, thus benefiting more from selective reuse.

\paragraph{Ablation on Reuse Patterns.}
We further compare different reuse scheduling strategies across block types, as shown in Tab.~\ref{tab: Ablation1}. In the \textit{``Split"} pattern, foreground-focused blocks are reused in early steps and background-focused blocks in later ones. In contrast, \textit{``Alternate"} reuse switches between them at each step segment. Split reuse provides stronger consistency across frames by stabilizing background modeling, while alternate reuse improves adaptability by interleaving static and dynamic feature attention. Notably, both are outperformed by the \textbf{Background-focused block reuse} strategy (\ie, reusing background-focused blocks with decreasing intervals), which aligns the reuse pattern with the generation trajectory. This method best exploits attention specialization across blocks, achieving the best trade-off between distortion (PSNR, SSIM) and perceptual realism (FID, LPIPS) with modest latency.

\paragraph{Ablation on Step-Level Reuse.}
To explore how reuse frequency affects performance, we vary the reuse interval, defined as the number of steps a cached block is retained before recomputation. As shown in Tab.~\ref{tab: Ablation0}, a \textit{``Stepwise"} strategy—where reuse intervals gradually decrease—consistently improves perceptual quality over fixed (\textit{``Step Average"}) or increasing (\textit{``Step Inverse"}) schedules. For instance, the \textbf{Background+Stepwise} strategy outperforms its inverse counterpart in both PSNR (26.44 vs. 24.49) and LPIPS (0.1203 vs. 0.2529), while being nearly twice as fast as the baseline (932 s vs. 1745 s). These findings highlight that reusing stable (background-focused) blocks early, followed by more frequent recomputation of dynamic (foreground-focused) blocks later, helps avoid stale information accumulation while maintaining quality.

\subsection{Ablation about resolution and GPU numbers}
We further evaluate our method at different resolutions (720p and 360p) and frame counts (129, 69, and 49 frames), using 1, 2, 4, and 8 GPUs to measure speedup over the original method. As shown in Fig.~\ref{fig: resolutions and video lengths}, our approach consistently outperforms the baseline. At 720p with 129 frames, we achieve a 1.87× speedup on a single GPU, maintaining 1.84× with 2 GPUs and 1.80× with 4 GPUs. However, with 8 GPUs, the speedup drops to 1.61× due to communication overhead. A similar pattern is observed for shorter videos, with diminishing efficiency as GPU count increases. These results demonstrate that our method effectively reduces computational time, particularly for high-resolution and long-duration, while maintaining robust scalability across multiple GPUs.

\subsection{Discussion on Multiple Objects}

To investigate the applicability of our attentional analysis in multi-object scenarios, we visualize the attention for a video with the prompt \textit{``a marathon with tens of runners"}, as shown in Fig.~\ref{fig: multi-onject}. Despite the presence of numerous objects, their significant motion and prominence categorize them as foreground, resulting in higher attention scores. The results demonstrate that the model maintains its ability to distinguish between background and foreground using specific blocks (\eg, blocks 12, 21, and 37), consistent with the single-object case.

%% file: sec/5_conclusion.tex
\section{Conclusion}

We conduct a systematic analysis of attention dynamics in the DiT framework and propose ProfilingDiT, a semantically guided caching strategy. Leveraging foreground-background segmentation and attention distribution priors, ProfilingDiT adaptively caches informative blocks and adjusts caching frequency across denoising steps. Experiments demonstrate that ProfilingDiT achieves significant acceleration with minimal degradation in video quality, highlighting its effectiveness for optimizing DiT-based diffusion models.
\paragraph{Future Work.} We plan to extend and deploy it across a broader range of video diffusion models to prove its generalizability and further enhance its robustness in diverse scenarios.